\documentclass[10pt, conference, compsocconf]{IEEEtran}

\usepackage{times}
\usepackage{helvet}
\usepackage{courier}
\usepackage{subfigure}
\usepackage{graphicx}
\usepackage{amsmath}
\usepackage{multirow}
\usepackage{comment}
\usepackage{amssymb}
\usepackage{multirow}
\usepackage{color}
\usepackage{algorithm}
\usepackage{algorithmic}
\usepackage{graphics}
\usepackage{rotating}

\begin{document}

\title{Representations and Ensemble Methods for Dynamic Relational Classification}

\author{\IEEEauthorblockN{Ryan A. Rossi}
\IEEEauthorblockA{Department of Computer Science\\
Purdue University\\
rrossi@cs.purdue.edu}
\and
\IEEEauthorblockN{Jennifer Neville}
\IEEEauthorblockA{Department of Computer Science\\
Purdue University\\
neville@cs.purdue.edu}}

\maketitle
\begin{abstract}
Temporal networks are ubiquitous and evolve over time by the addition, deletion, and changing of links, nodes, and attributes. Although many relational datasets contain temporal information, the majority of existing techniques in relational learning focus on static snapshots and ignore the temporal dynamics. We propose a framework for discovering temporal representations of relational data to increase the accuracy of statistical relational learning algorithms. 
The temporal relational representations serve as a basis for classification, ensembles, and pattern mining in evolving domains.
The framework includes (1) selecting the time-varying relational components (links, attributes, nodes), (2) selecting the temporal granularity (i.e., set of timesteps), (3) predicting the temporal influence of each time-varying relational component, and (4) choosing the weighted relational classifier. 
Additionally, we propose temporal ensemble methods that exploit the temporal-dimension of relational data. These ensembles outperform traditional and more sophisticated relational ensembles while avoiding the issue of learning the most optimal representation. 
Finally, the space of temporal-relational models are evaluated using a sample of classifiers. In all cases, the proposed temporal-relational classifiers outperform competing models that ignore the temporal information. The results demonstrate the capability and necessity of the temporal-relational representations for classification, ensembles, and for mining temporal datasets.
\end{abstract}


\IEEEpeerreviewmaketitle

\newcommand\TT{\rule{0pt}{2.3ex}}
\newcommand\BB{\rule[-1.0ex]{0pt}{0pt}}
\newcommand\T{\rule{0pt}{3.8ex}}
\newcommand\B{\rule[-1.8ex]{0pt}{0pt}}

\section{Introduction}
Temporal-relational information is seemingly ubiquitous; it is present in domains such as the Internet, citation and collaboration networks, communication/email networks, social networks, biological networks, among many others. These domains all have attributes, links, and/or nodes changing over time which are important to model.
We conjecture that discovering an accurate \textit{temporal-relational representation} disambiguates the true nature and strength of links, attributes, and nodes.
However, the majority of research in relational learning has focused on modeling static snapshots~\cite{chakrabarti:98,domingos:01,neville:kdd05} and has largely ignored the utility of learning and incorporating temporal dynamics into relational representations.

Temporal relational data has three main components (i.e., attributes, nodes, links) that vary in time. First, the attribute values might change over time (e.g., research area of an author). Secondly, links might be created and deleted throughout time (e.g., friendships or a paper citing a previous paper). Thirdly,  nodes might be activated and deactivated throughout time (e.g., a person might not send an email for a few days). Additionally, in a \textit{temporal prediction task}, the attribute to predict is changing throughout time (e.g., predicting a network anomaly) whereas in a static prediction task the predictive attribute remains constant.

Consequently, the space of temporal relational models is defined by considering the set of relational elements that might change over time such as the attributes, links, and nodes. Additionally, the space of temporal-relational representations depends on a \textit{temporal weighting} and the \textit{temporal granularity}.  The temporal weighting attempts to predict the influence of the links, attributes and nodes by decaying the weights of each with respect to time whereas the temporal granularity restricts links, attributes, and nodes with respect to some window of time. The most optimal temporal-relational representation and the corresponding temporal classifier depends on the particular temporal dynamics of the links, attributes, and nodes present in the data and also on the domain and type of network (e.g., social networks, biological networks).

In this work, we address the problem of selecting the most optimal temporal-relational representation to increase accuracy of predictive models. The space of \textit{temporal-relational representations} leads us to propose the \textbf{(1)} temporal-relational classification framework and \textbf{(2)} temporal ensemble methods (e.g., temporally sampling, randomizing, and transforming features) that leverage time-varying links, attributes, and nodes. We evaluate these temporal-relational models on a variety of classification tasks and evaluate each under various constraints. Finally, we explore the utility of the framework for \textbf{(3)} mining temporal datasets and discovering temporal patterns. The results demonstrate the importance and scalability of the temporal-relational representations for classification, ensembles, and for mining temporal datasets.

\section{Related Work}
Most previous work uses static snapshots or significantly limits the amount of temporal information used for relational learning.
Sharan et. al.~\cite{sharan:icdm08} assumes a strict temporal-representation that uses kernel estimation for links and includes these into a classifier. They do not consider multiple temporal granularities (all information is used, statically) and the attributes and nodes are not weighted. In addition, they focus only on one specific temporal pattern and ignore the rest whereas we explore many temporal-relational representations and propose a flexible framework capable of capturing the temporal patterns of links, attributes, and nodes. Moreover, they only evaluate and consider static prediction tasks. Other work has focused on discovering temporal patterns between attributes~\cite{mei:05}. There are also temporal centrality measures that capture properties of the network structure~\cite{tang2010analysing}.

\section{Temporal-Relational Classification Framework}
The temporal-relational classification framework is defined with respect to the possible transformations of links, attributes, or nodes (as a function of time). The temporal weighting (e.g., exponential decay of past information) and temporal granularity (e.g., window of timesteps) of the links, attributes and nodes form the basis for any arbitrary transformation with respect to the temporal information (See Table~\ref{table:trcf}). The discovered temporal-relational representation can be applied for mining temporal patterns, classification, and as a means for constructing temporal-ensembles. An overview of the temporal-relational representation discovery is provided below:

\begin{enumerate}
\item For each \textsc{Relational Component}
\begin{list}{$-$}{}
\item Links, Attributes, or Nodes
\end{list}{}{}
\item Select the \textsc{Temporal Granularity}
\begin{list}{$\star$}{}
\item Timestep\;\;$t_i$
\item Window   \;$\{t_i, t_{i+1}...,t_j\}$
\item Union \;\;\;\;$T = \{t_0,...,t_n\}$
\end{list}{}{}
\item Select the \textsc{Temporal Influence}
\begin{list}{$\star$}{}
\item Weighted
\item Uniform
\end{list}{}{}
\footnotesize Repeat steps 1-3 for each relational component. \normalsize
\item Select the Modified \textsc{Relational Classifier}
\begin{list}{$\star$}{}
\item Relational Bayes Classifier (RBC)
\item Relational Probability Trees (RPT)
\end{list}{}{}
\end{enumerate}

Table~\ref{table:trcf} provides an intuitive view of the possible temporal-relational representations. For instance, the recent TVRC model is a special case of the proposed framework where the links, attributes, and nodes are unioned and the links are weighted.

\subsection{Relational Components: Links, Attributes, Nodes}
The data is represented as an attributed graph $D=(G,\mathbf{X})$. The graph $G=(V,E)$ represents a set of $N$ nodes, such that $v_i \in V$ corresponds to node $i$ and each edge $e_{ij} \in E$ corresponds to a link (e.g., email) between nodes $i$ and $j$. The attribute set:
$$\mathbf{X} = 
\left(
\begin{array}{ll} 
 \mathbf{X^V}=[X^1,X^2, ... ,X^{m_v}], \\
 \mathbf{X^E}=[X^{m_v+1},X^{m_v+2}, ... ,X^{m_v+m_e}]
\end{array}
\right)$$
\noindent may contain observed attributes on both the nodes ($\mathbf{X^V}$) and the edges ($\mathbf{X^E}$). Below we use $X^m$ to refer to the generic $m^{th}$ attribute on either nodes or edges.

There are three aspects of relational data that may vary over time. First, the values of attribute $X^m$ may vary over time.

Second, edges may vary over time. This results in a different data graph $G_t=(V,E_t)$ for each time step $t$, where the nodes remain constant but the edge set may vary (i.e., $E_{t_i} \neq E_{t_j}$ for some $i, j$). Third, a nodes existence may vary over time (i.e., objects may be added or deleted). This is also represented as a set of data graphs $G'_t=(V_t,E_t)$, but in this case both the nodes and edge sets may vary.
Let $D_t=(G_t,\mathbf{X}_t)$ refer to the dataset set at time $t$, where  $G_t=(V,E_t,W^E_t)$ and $\mathbf{X}_t=(\mathbf{X}^V_{t},\mathbf{X}^E_{t}, W^X_t)$. Here $W_t$ refers to a function that assigns weights on the edges and attributes that are used in the classifiers below. We define $W^E_t(i,j)=1$ if $e_{ij} \in E_t$ and $0$ otherwise. Similarly, we define $W^X_t(x^m_i)=1$ if $X^m_i=x^m_i \in \mathbf{X^m_t}$ and $0$ otherwise.

\begin{table}[t!]
\caption{\textsc{\textbf{Temporal-Relational Representation.}}}
\label{table:trcf}
\begin{center}
\begin{small}
\begin{tabular}{ r || c | c | c || c | c | c ||}
\multicolumn{1}{c||}{\TT \BB  \textbf{}} &  \multicolumn{3}{c||}{\TT \BB  \textbf{\textsc{Uniform}}} & \multicolumn{3}{c||}{\TT \BB  \textbf{\textsc{Weighting}}} \\

\TT \BB \rotatebox{90}{\textbf{\textsc{}}} & \rotatebox{90}{\textbf{Timestep}} & \rotatebox{90}{\textbf{Window}} & \rotatebox{90}{\textbf{Union}} & \rotatebox{90}{\textbf{Timestep}} & \rotatebox{90}{\textbf{Window}} & \rotatebox{90}{\textbf{Union}} \\
\hline
\hline
\TT \BB \textbf{Edges} &  &  &  &  &  &    \\
\hline
\TT \BB \textbf{Attributes} &  &  &  &  &  &    \\
\hline
\TT \BB \textbf{Nodes} &  &  &  &  &  &    \\
\hline
\hline
\end{tabular}
\end{small}
\end{center}
\end{table}

\subsection{Temporal Granularity}
Traditionally, relational classifiers have attempted to use all the data~\cite{sharan:icdm08}. Conversely, the appropriate temporal granularity (i.e., set of timesteps) can be learned to improve classification accuracy.  We briefly define the three general classes evaluated in this work for varying the temporal granularity of the links, attributes, and nodes.

\begin{enumerate}
\item \noindent \textbf{Timestep.} The timestep models only use a single timestep $t_i$ for learning.

\item \noindent \textbf{Window.}  The window models use a sliding window of timesteps $\{t_i, t_{i+1}...,t_j\}$ for learning. The space of window models is by far the largest.

\item \noindent \textbf{Union.} The union model uses all the previous temporal information for learning.
\end{enumerate}

The timestep and union models are separated into distinct classes for clarity in evaluation and for pattern mining.

\subsection{Temporal Influence: Links, Attributes, Nodes}
The influence of the relational components over time are predicted using temporal weighting. The temporal weights can be viewed as probabilities that a relational component is still active at the current time step $t$, given that it was observed at time $(t-k)$.
Conversely, the temporal influence of a relational component might be treated uniformly.
Additionally, weighting functions can be chosen for different relational components with varying temporal granularities. 
For instance, the temporal influence of the links might be predicted using the exponential kernel while the attributes are uniformly weighted but have a different temporal granularity than the links. 

\begin{enumerate}
\item \noindent \textbf{Weighting.} We investigated three temporal weighting functions:

\begin{itemize}
\item \textit{Exponential Kernel.} The exponential kernel weights the recent past highly and decays the weight rapidly as time passes~\cite{cortes:01}. The kernel function $K_E$ for temporal data is defined as:
\begin{displaymath}
K_E(D_i;t,\theta) = (1 - \theta)^{t-i}\theta W_{i}
\end{displaymath}

\item \textit{Linear Kernel.} The linear kernel decays more gently and retains the historical information longer. The linear kernel $K_L$ for the data is defined as:
\begin{displaymath}
K_L(D_i;t,\theta) = \theta W_{i} (\frac{t_{*} - t_i + 1}{t_{*} - t_o + 1})
\end{displaymath}

\item \textit{Inverse Linear Kernel.} The inverse linear kernel $K_{IL}$ lies between the exponential and linear kernels when moderating the contribution of historical information. The inverse linear kernel for the data is defined as:
\begin{displaymath}
K_{IL}(D_i;t,\theta) = \theta W_{i} (\frac{1}{t_i - t_{o} + 1})
\end{displaymath}
\end{itemize}

\item \noindent \textbf{Uniform.} The relational component(s) could be assigned uniform weights across time for the selected temporal granularity (e.g., traditional classifiers assign uniform weights, but they do not select the appropriate temporal granularity).
\end{enumerate}

\subsection{Temporal-Relational Classification}
Once the temporal granularity and the temporal weighting are selected for each relational component, then a temporal-relational classifier is learned.
Modified versions of the RBC~\cite{neville:icdm03} and the RPT~\cite{neville:kdd03} are applied with the temporal-relational representation. However, any relational model that has been modified for weights is suitable for this phase. We extended RBCs and RPTs since they are interpretable, diverse, simple, and efficient. We use $k$-fold cross-validation to learn the ``best'' model. Both classifiers are extended for learning and inference through time.

\textbf{Weighted Relational Bayes Classifier.}
RBCs extend naive Bayes classifiers to relational settings by
treating heterogeneous relational subgraphs as a homogenous set of
attribute multisets. For example, when modeling the dependencies between the topic of a paper and the topics of its references, the topics of those references form
multisets of varying size (e.g., \{NN, GA\}, \{NN, NN, RL, NN, GA\}). The RBC models these heterogenous multisets by assuming that each value of the multiset is
independently drawn from the same multinomial distribution. This
approach is designed to mirror the independence assumption of the
naive Bayesian classifier~\cite{domingos:97}. In addition to the
conventional assumption of attribute independence, the RBC also
assumes attribute value independence within each multiset. More formally, for a class label $C$, attributes $\mathbf{X}$, and related items $R$, the RBC calculates the probability of $C$ for an item $i$ of type $G(i)$ as follows:
\begin{equation*} \label{orbc}
P(C^i | \mathbf{X}, R) \propto \!\!\!\!\!\! \prod_{X_m \in \mathbf{X^{G(i)}}} \!\!\!\!\!\! P(X_m^i | C) \;  \prod_{j \in R} \prod_{X_k \in \mathbf{X^{G(j)}}} \!\!\!\!\!\! P(X^j_k | C)  P(C) 
\end{equation*}

\begin{figure}[b!]
\centering
 \includegraphics[width=3.2in]{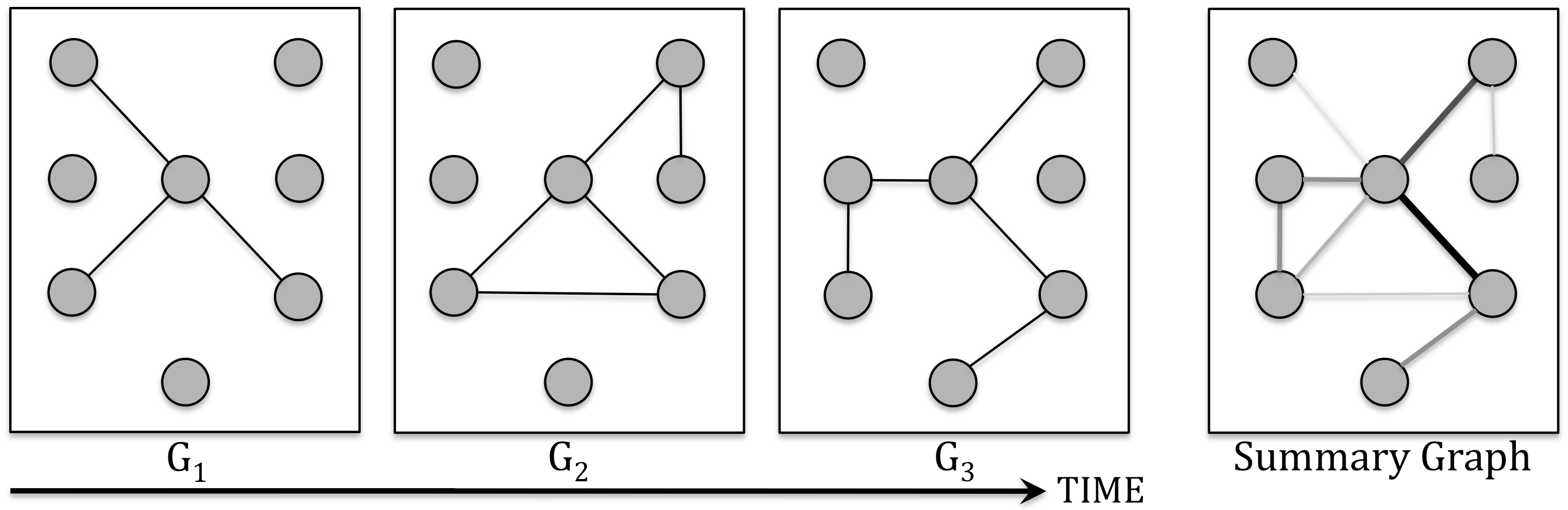}
 \vspace{-2.0mm}
\caption{Temporal Link Weighting}
\label{fig:summary-graph}
\end{figure}

\begin{figure}[b!]
\centering
 \includegraphics[width=3.2in]{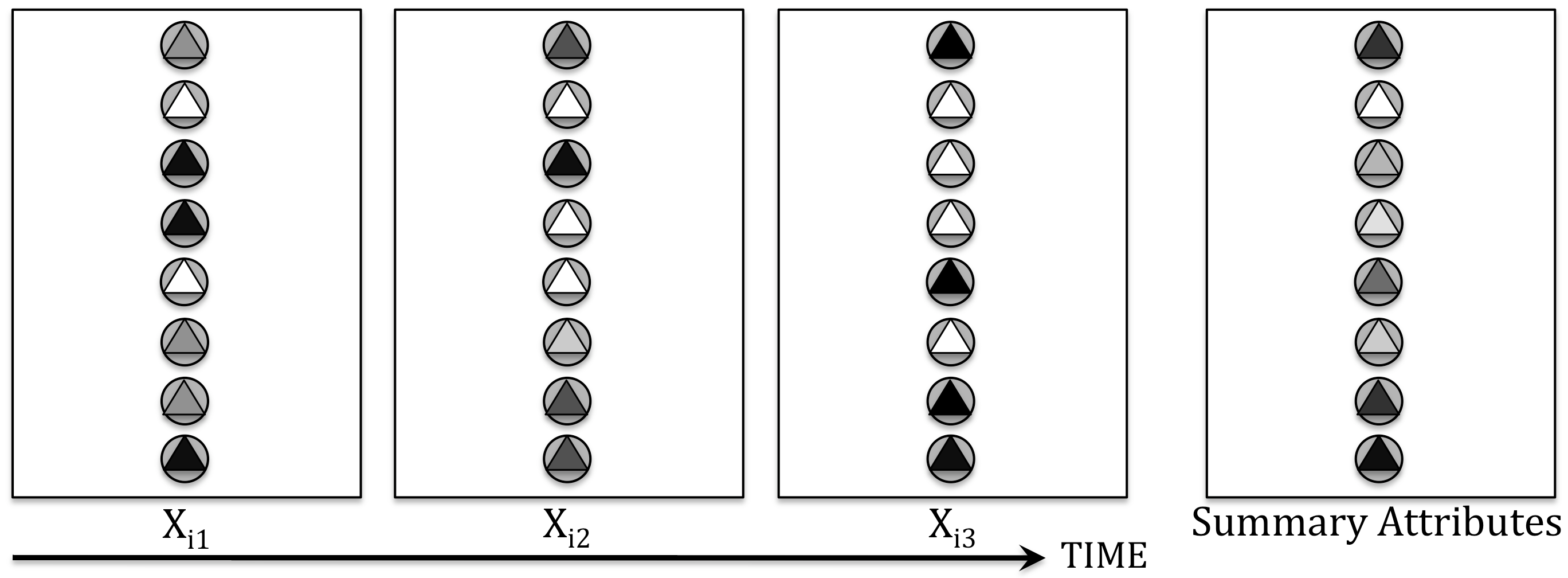}
  \vspace{-4.0mm}
\caption{Temporal Attribute Weighting} 
\label{fig:summary-attributes}
\end{figure}

\begin{figure}[b!]
\centering
 \includegraphics[width=3.2in]{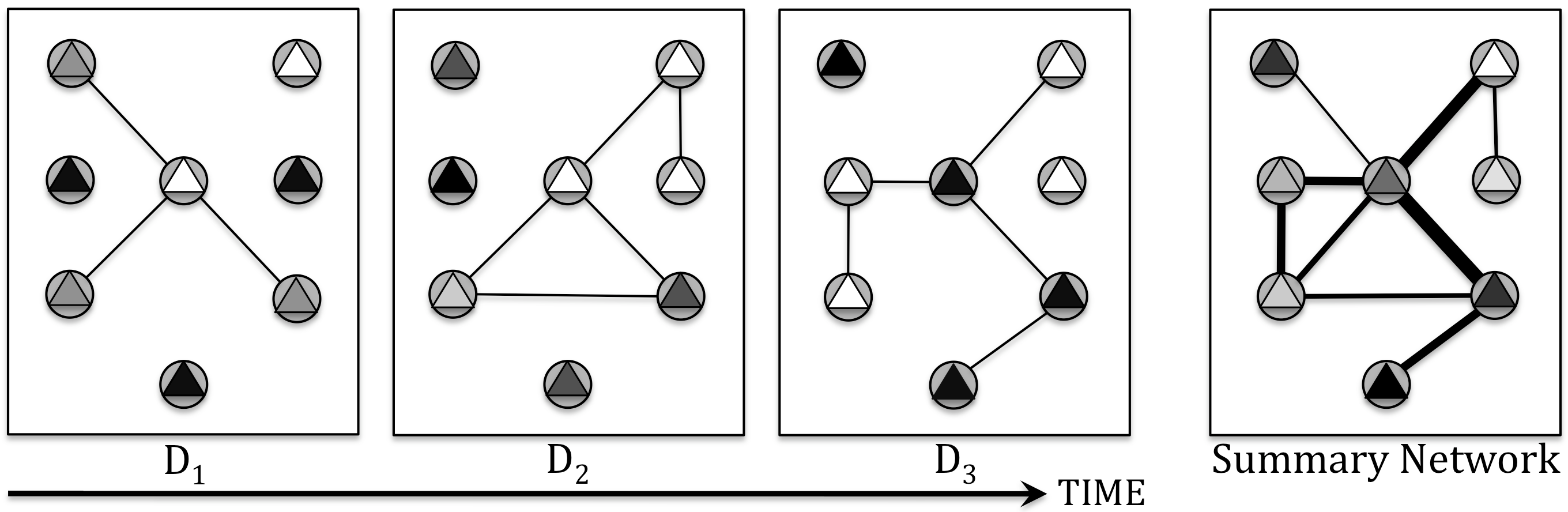}
  \vspace{-4.0mm}
\caption{Graph and Attribute Weighting} 
\label{fig:summary-data}
\end{figure}
\vspace{-2.0mm}

\begin{figure}[t!]
    \subfigure[Links weighting]
    	{\label{fig:tvrc-model}\includegraphics[width=1.6in]{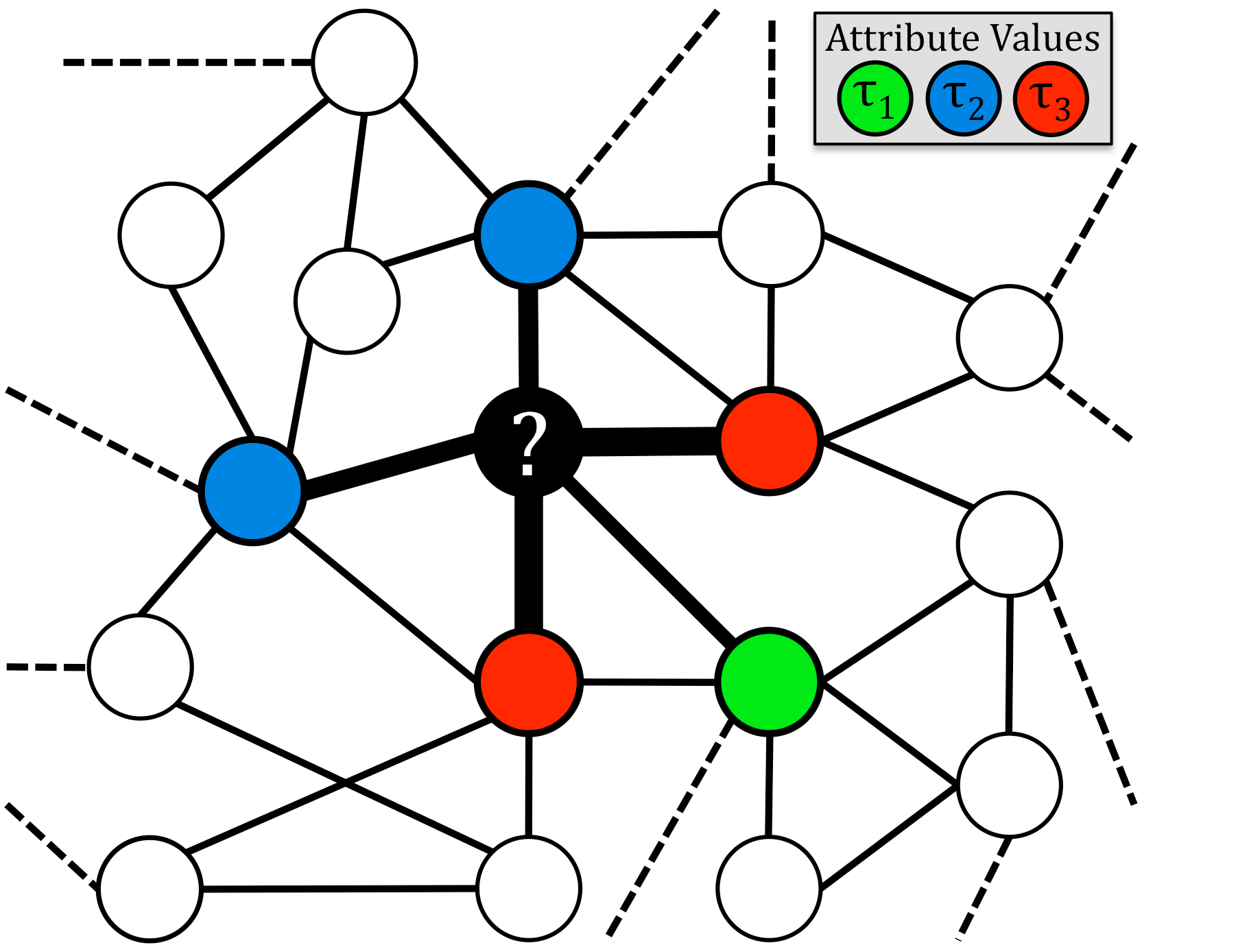}}
    \subfigure[Link and attribute weighting]
    	{\label{fig:tenc-model}\includegraphics[width=1.6in]{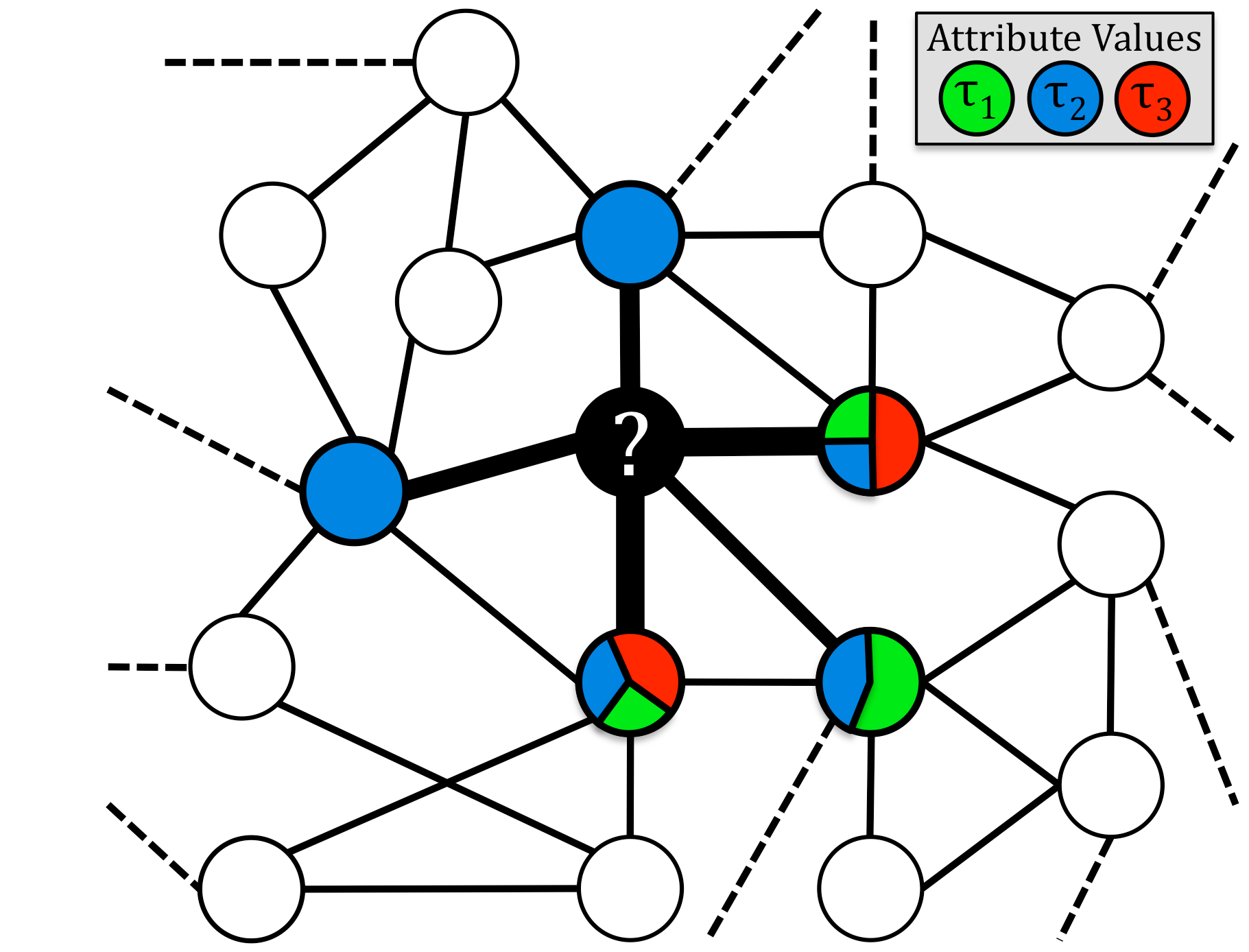}}
  \caption{\textbf{(a)} The feature calculation that includes only the temporal link weights.  \textbf{(b)} The feature calculation that incorporates \textit{both} the temporal attribute weights and the temporal link weights.}
  \label{fig:models}
 \vspace{-6.mm}
\end{figure}

\textbf{Weighted Relational Probability Trees.}
RPTs extend standard probability estimation trees to a relational
setting in which data instances are heterogeneous and interdependent.
The algorithm for learning the structure and parameters
of a RPT searches over a space of relational features that
use aggregation functions (e.g. AVERAGE, MODE, COUNT) to dynamically
propositionalize relational data multisets and create binary
splits within the RPT.

\textit{Learning.}
The RBC uses standard maximum likelihood learning with Laplace correction for zero-values. More specifically, the sufficient statistics for each conditional probability distribution are computed as weighted sums of counts based on the link and attribute weights. The RPT uses the standard RPT learning algorithm except that the aggregate functions are computed after the appropriate links and attributes weights are included with respect to the selected temporal granularity (shown in Figure~\ref{fig:models}).

\textit{Prediction.}
For prediction we compute the summary data $D_{S_t}$ at time t $-$ the time step for which the model is being applied. The learned model for time $(t-1)$ to $D_{S_t}$. The weighted classifier is appropriately augmented to incorporate the weights for $D_{S_t}$.

\section{Temporal Ensemble Methods}
Ensemble methods have traditionally been used to improve predictions by considering a weighted vote from a set of classifiers~\cite{dietterich2000ensemble}. We propose temporal ensemble methods that exploit the \textit{temporal dimension of relational data} to construct more accurate predictors. This is in contrast to traditional ensembles that disregard the temporal information. The \textit{temporal-relational classification framework} and in particular the temporal-relational representations of the time-varying links, nodes, and attributes form the basis of the temporal ensembles (i.e., used as a wrapper over our framework). The proposed temporal ensemble techniques are assigned to one of the five methodologies described below.

\subsection{Transforming the Temporal Nodes and Links}
The first temporal-ensemble method learns a set of classifiers where each of the classifiers are applied after the link and nodes are sampled from each discrete timestep according to some probability. This sampling strategy is performed after constructing the temporal-relational representation where the temporal weighting and temporal granularity have been selected. Additionally, the sampling probabilities for each timestep can be chosen to be biased toward the present or the past. In contrast to applying a sampling strategy across time, we might transform the time-varying nodes and links using the methods described in the framework.

\subsection{Sampling or Transforming the Temporal Feature Space}
The second type of temporal ensemble method transforms the temporal feature space by localizing randomization (for attributes at each timestep), weighting, or by varying the temporal granularity of the features. Additionally, we might use only one temporal weighting function but learn different decay parameters or resample from the temporal features. The temporal features could also be clustered (using varying decay parameter or randomizations), similar to the dynamic topic discovery models evaluated later in the paper.

\subsection{Adding Noise or Randomness}
A temporal ensemble based on adding noise along the temporal dimension of the data may significantly increase generalization and performance. Suppose, we randomly permute the nodes feature values across the timesteps (i.e., a nodes recent behavior is observed in the past and vice versa) or links between nodes are permuted across time.

\subsection{Transforming the Time-Varying Class Labels}
These temporal ensemble methods introduce variance in the classifiers by randomly permuting the previously learned labels at $t$-$1$ (or more distant) with the the true labels at $t$.

\subsection{Multiple Classification Algorithms and Weightings}
A temporal ensemble may be constructed by randomly selecting from a set of classification algorithms (i.e., RPT, RBC, wvRN, RDN), while using equivalent temporal-relational representations or by varying the representation with respect to the temporal weighting or granularity. Notably, an ensemble using RPT and RBC significantly increases accuracy, most likely due to the diversity of these temporal classifiers (i.e., correctly predicting different instances). Additionally, the temporal-classifiers might be assigned weights based on cross-validation (or Bayesian approach).

\section{Methodology}
We describe the datasets and define a few representative temporal-relational classifiers from the framework.

\subsection{Datasets}
For evaluating the framework, we use a range of both static (i.e., prediction attribute is constant as a function of time) and temporal prediction tasks (i.e., prediction attribute changes between timesteps).
 
\textbf{\textsc{PyComm} Developer Communication Network.}
We analyze email and bug communication networks extracted from the Python development environment (www.python.org). We use the python-dev mailing list archive for the period 01/01/07$-$09/30/08. The sample contains 13181 email messages, among 1914 users. Bug reports were also collected and we constructed a second \emph{bug} discussion network. The sample contained 69435 bug comments among 5108 users. The size of the timesteps are three months.

We also extracted text from emails and bug messages and use it to dynamically model the topics between individuals and teams. Additionally, we discover temporal centrality attributes (i.e., clustering coefficient, betweenness). The prediction task is whether a developer is effective (i.e., if a user closed a bug in that timestep).

\begin {table}[!h]
\caption{Generated attributes from the \textsc{PyComm} Network}
\centering
\begin{small}
\begin{tabular}{c||ll} \hline
\multicolumn{3}{c}{\textbf{\TT \BB Python Communication Network Attributes}} \\ \hline\hline

\TT \BB \multirow{9}{*}{} & Conv Tool & Build  \\
& Demos \& tools & Dist Utils  \\
& Documentation & Doc Tools  \\
\textbf{Team} & Installation & InterpCore  \\
\textbf{Membership} & Regular Expr & Tests  \\
& Unicode & Windows  \\
& Ctypes & Ext Modules \\
& Idle & LibraryLib \\
& Tkinter & XML \\ \hline
\TT \BB \multirow{1}{*}{\textbf{Performance}} & Assigned To  & \textsc{[Has Closed]}  \\ \hline

\TT \BB \multirow{2}{*}{} \textbf{Communication} & Comm. Count & Bug Comm. \\
\textbf{Attributes} & Email Comm. &  \\ \hline

\TT \BB \multirow{2}{*}{\textbf{User Topics}} & Topic & Email Topic \\ 
& Bug Topic & \\ \hline

\TT \BB \multirow{2}{*}{}\textbf{Temporal} & Eigenvector &  Cluster. Coeff.\\
\textbf{Centrality}& Betweenness &  Degree\\ \hline

\TT \BB \multirow{3}{*}{} & Edge Count & Edge Topic  \\
\textbf{Link Attributes} & Email Count  & Email Topic  \\
 & Bug Count & Bug Topic\\ \hline

\end{tabular}
	\label{tab:Relational and Intrinsic Attributes}
\end{small}
\end{table}

\textbf{\textsc{Cora} Citation Network.}
The \textsc{Cora} database contains authorship and citation information about CS research papers extracted automatically from the web. The prediction tasks are to predict one of seven machine learning papers and to predict AI papers given the topic of its references. In addition, these techniques are evaluated using the most prevalent topics its authors are working on through collaborations with other authors.

\subsection{Temporal Models}
The space of temporal-relational models are evaluated using a representative sample of classifiers with varying temporal-relational weightings and granularities. For every timestep $t$, we learn a model on $D_{t}$ (i.e., some set of timesteps) and apply the model to $D_{t+1}$. The utility of the temporal-relational classifiers and representation are measured using the area under the ROC curve (AUC). Below, we briefly describe a few classes of models that were evaluated.

\begin{itemize}
\item \textbf{TENC}: The TENC models predict the temporal influence of both the links and attributes.
\item \textbf{TVRC}: This model weights only the links using all previous timesteps.
\item \textbf{Union Model}: The union model uses all links and nodes up to and including $t$ for learning. 
\item \textbf{Window Model}: The window model uses the data $D_{t-1}$ for prediction on $D_{t}$ (unless otherwise specified).
\end{itemize}

We also compare simpler models such as the RPT (relational information only) and the DT (non-relational) that ignore any temporal information. Additionally, we explore many other models, including the class of window models, various weighting functions (besides exponential kernel), and built models that vary the set of windows in TENC and TVRC.

\section{Experiments} \label{sec:experiments}
We first evaluate temporal-relational representations for improving classification models. These models are evaluated using different types of attributes (e.g., relational only vs. non-relational) and also by using different types of discovered attributes (e.g., temporal centrality, team attributes, communication). The results demonstrate the utility of the temporal-relational classifiers, their representation, and the discovered temporal attributes. We also identify the minimum temporal information (i.e., simplest model) required to outperform classifiers that ignore the temporal dynamics. 
Furthermore, the proposed temporal ensemble methods (i.e., temporally sampling, randomizing, and transforming features) are evaluated and the results demonstrate significant improvements over traditional and relational ensemble methods.

We then focus on models that vary the temporal-granularity and apply these for mining temporal patterns and more generally for discovering the nature of the time-varying links and attributes. 
Finally, we apply temporal textual analysis, generate topic features, and annotate the links and nodes with their corresponding topics over time.
The significance of the evolutionary topic patterns are evaluated using a classification task. The results indicate the effectiveness of the temporal textual analysis for discovering time-varying features and incorporating these patterns to increase the accuracy of a classification task. For brevity, we omit many plots and comparisons.

\begin{figure}[t!]
\centering
 \includegraphics[width=2.6in]{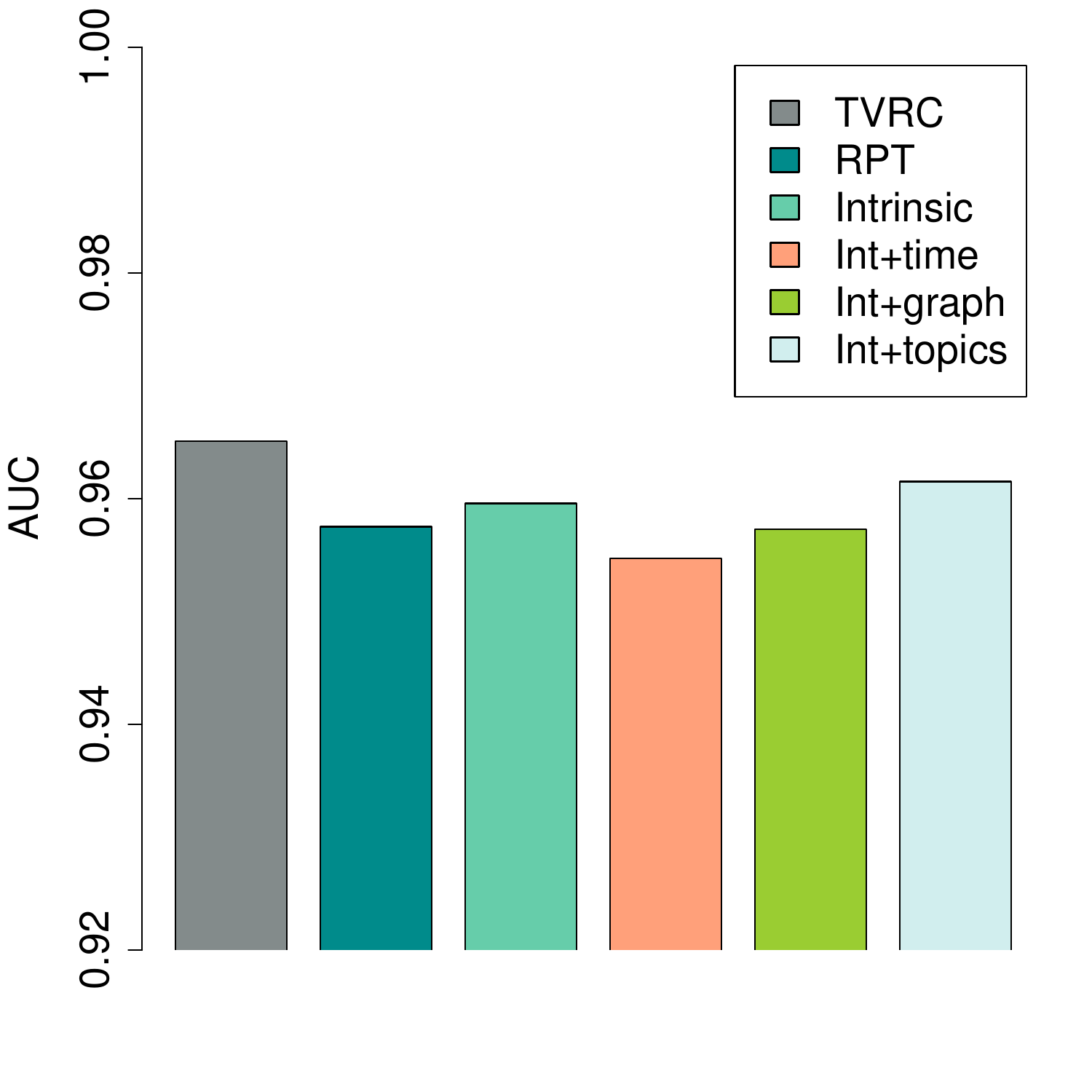}
 \vspace{-4.mm}
\caption{We compare a primitive \textit{temporal} model (TVRC) to competing relational (RPT), and non-relational (DT) models. The AUC is averaged across the timesteps.}
\label{single-results}
 \vspace{-6.mm}
\end{figure}

\subsection{Single Models}
We provide examples of temporal-relational models from the proposed framework and show that in all cases the performance of classification improves when the temporal dynamics are appropriately modeled. 

\noindent\textbf{Temporal, Relational, and Non-relational Information.}
We first assess the utility of the temporal, relational, and non-relational information. 
In particular, we are interested in this information as it pertains to the construction of features and their selection and pruning from the model. 
For these experiments, we compare the most primitive models such as TVRC (i.e., uses temporal-relational information), RPT (i.e., only relational information), and a decision tree that uses only non-relational information. Additionally, we learn these models using various types of attributes and explore the utility of each with respect to the temporal, relational or non-relational information.

Figure~\ref{single-results} compares TVRC (i.e., a primitive temporal-relational classifier) with the RPT and DT models that use more features but ignore the temporal dynamics of the data. We find the TVRC to be the simplest temporal-relational classifier that still outperforms the others. Interestingly, the discovered topic features are the only additional features that improve performance of the DT model. This is significant as these attributes are discovered by dynamically modeling the topics, but are included in the DT model as simple non-relational features (i.e., no temporal weighting or granularity, ...). We also find that in some cases the selective learner chooses a suboptimal feature when additional features are included in the basic DT model (see Figure~\ref{single-results}). More surprisingly, the base RPT model does not improve performance over the DT model, indicating the significance of moderating the relational information \textit{with} the temporal dynamics.

\begin{figure}[t!]
\vspace{-6.mm}
\centering
 \includegraphics[width=3in]{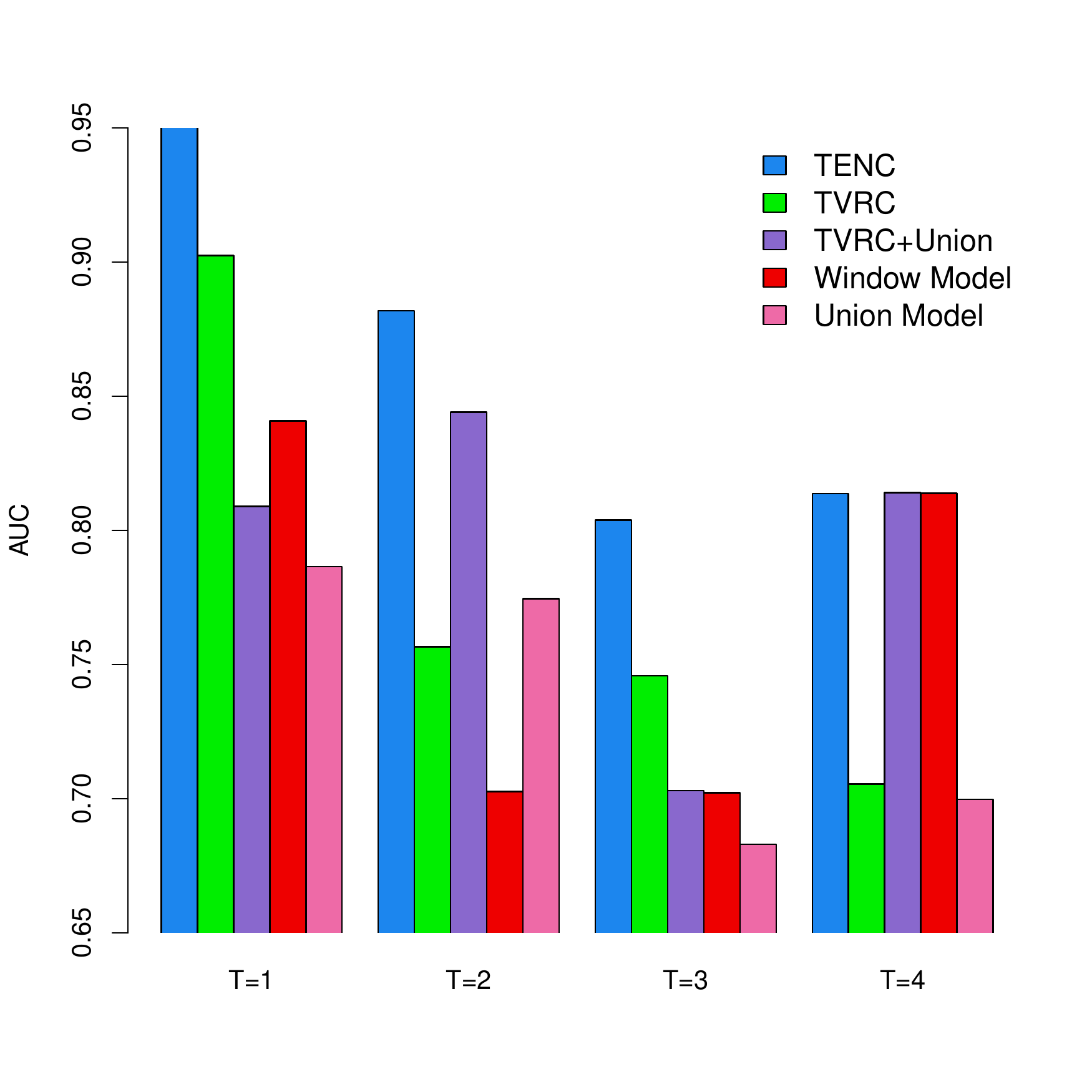}
 \vspace{-8.mm}
\caption{Exploring the space of temporal-relational models. We evaluate significantly different temporal-relational representations from the proposed framework. This experiment uses the \textsc{PyComm} network, but focuses on time-varying relational attributes.}
\label{fig:pycomm-temporal-models}
\vspace{-6.mm}
\end{figure}

\noindent\textbf{Exploring Temporal-Relational Models.}
We focus on exploring a small but representative set of temporal-relational models from the proposed framework. To more appropriately evaluate their temporal-representations, we chose to remove highly correlated attributes (i.e., that are not necessarily temporal patterns, or motifs), such as assignedto in the \textsc{PyComm} prediction task. In Figure~\ref{fig:pycomm-temporal-models}, we find that TENC outperforms the other models over all timesteps. This proposed class of models is significantly more complex than TVRC (and most other models) since the temporal influence of both links and attributes are learned.

We then explored learning the appropriate temporal granularity but with respect to the TVRC model. Figure~\ref{fig:pycomm-temporal-models} shows the results from two models in the TVRC class where we attempt to tease apart the superiority of TENC (i.e., weighting or granularity). However, both models outperform one another on different timesteps, indicating the necessity for a more precise temporal-representation that optimizes the temporal granularity by selecting the appropriate decay parameters for links and attributes (i.e., in contrast to a more strict representation of including the links or not). The window and union models perform significantly worse, but are significantly more efficient and scalable for billion node temporal datasets while still including some temporal information based on the granularity of the links and attributes. Similar results were found using \textsc{Cora} and other base classifiers such as RBC.

We have also experimented searching over many temporal weighting functions and found the exponential decay to be the most appropriate for both links and attributes in the proposed prediction tasks. The most optimal temporal-relational representation depends on the temporal dynamics and nature of the network under consideration (e.g., social networks, biological networks, citation networks). Nevertheless, multiple temporal weightings and granularities are found to be useful for constructing robust temporal ensembles that significantly reduce error and variance (i.e., compared to single temporal-relational classifiers and more importantly relational and traditional ensembles).

The accuracy of classification generally increases as more temporal information is included in the representation. However, this may lead to overfitting or other biases. On the other hand, the more complex temporal-relational representations aid in the mining of temporal patterns. For instance, the use of the evolutionary topic patterns for improving classification by moderating both the links and attributes over time (See Section~\ref{sec:dynamic-topics}).

\noindent \textbf{Selective Temporal Learning.}
We also explored ``selective temporal learning" that uses multiple temporal weighting functions (i.e., and temporal granularities) for the links and attributes. The motivation for such an approach is that the influence of each temporal component should be modeled independently, since any two attributes (or links) are likely to decay at different rates. However, the complexity and the utility of the learned temporal-relational representation depends on the ability of the selective learner to select the best temporal features (derived from weighting or varying the temporal granularity of attributes and links) without overfitting or causing other problems. We found that the selective temporal learning performs best for simpler prediction tasks, however, it still frequently outperforms classifiers that ignore the temporal information.

\begin{figure}[b!]
 \vspace{-6.mm}
\centering
 \includegraphics[width=2.6in]{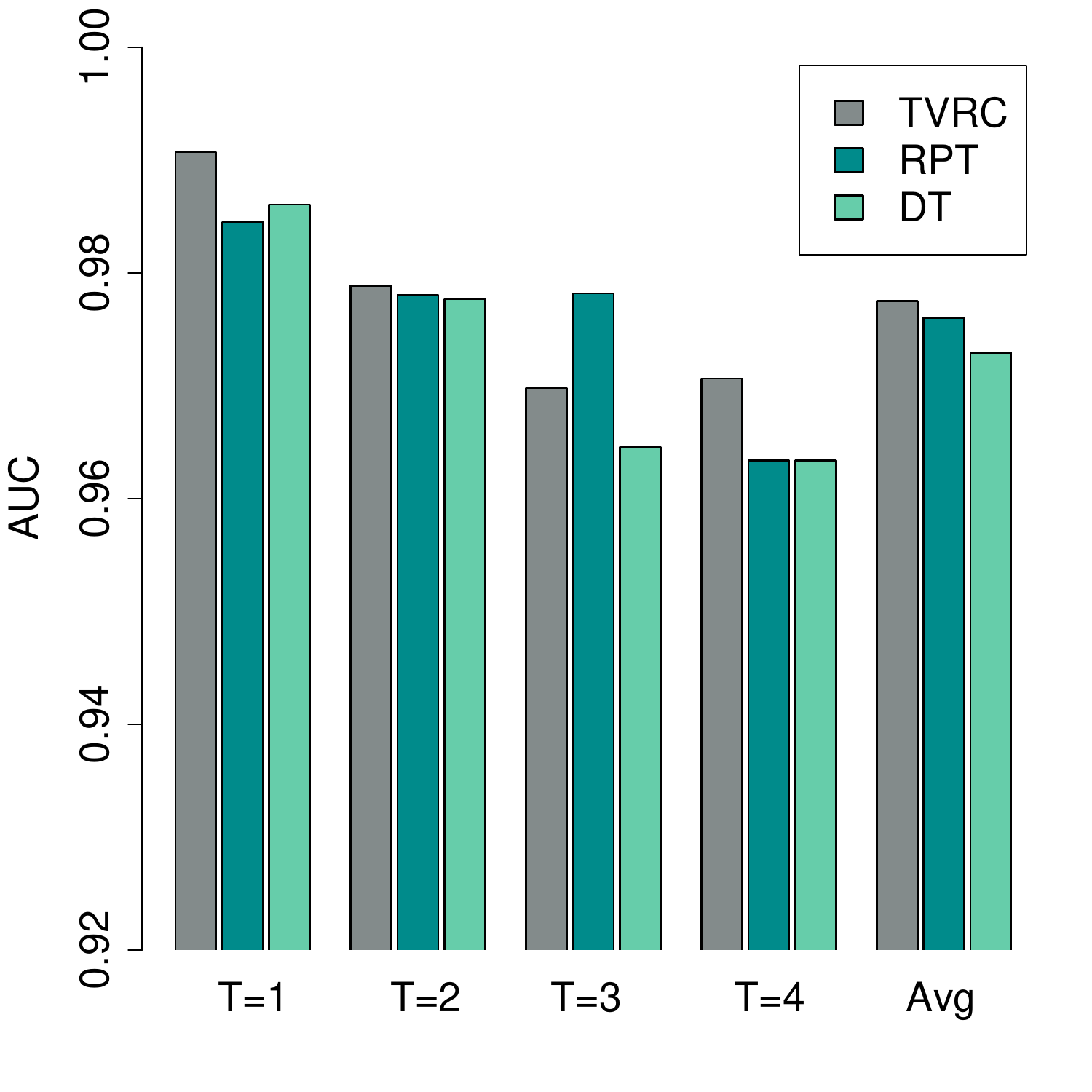}
 \vspace{-8.mm}
\caption{Comparing Temporal, Relational, and Traditional Ensembles}
\label{fig:forest-results}
\end{figure}

\subsection{Temporal-Ensemble Models}\label{sec:eval-temporal-ensembles}
Instead of directly learning the most optimal temporal-relational representation to increase the accuracy of classification, we use \textit{temporal ensembles} by varying the relational representation with respect to the temporal information. These ensemble models reduce error due to variance and allow us to assess which features are the most relevant to the domain with respect to the relational or temporal information.

\noindent \textbf{Temporal, Relational, and Traditional Ensembles.}
We first resampled the instances (nodes, links, features) repeatedly and then learn TVRC, RPT, and DT models. Across almost all the timesteps, we find the temporal-ensemble that uses various temporal-relational representations outperforms the relational-ensemble and the traditional ensemble (see Figure~\ref{fig:forest-results}). The temporal-ensemble outperforms the others even when a the minimum amount of temporal information is used (e.g., time-varying links). More sophisticated temporal-ensembles can be constructed to further increase accuracy. For instance, we have investigated ensembles that use significantly different temporal-relational representations (i.e., from a wider range of model classes) and ensembles that use various temporal weighting parameters. In all cases, these ensembles are more robust and increase the accuracy over more traditional ensemble techniques (and single classifiers).  

\begin{figure}[t!]
 \vspace{-6.mm}
\centering
 \includegraphics[width=2.6in]{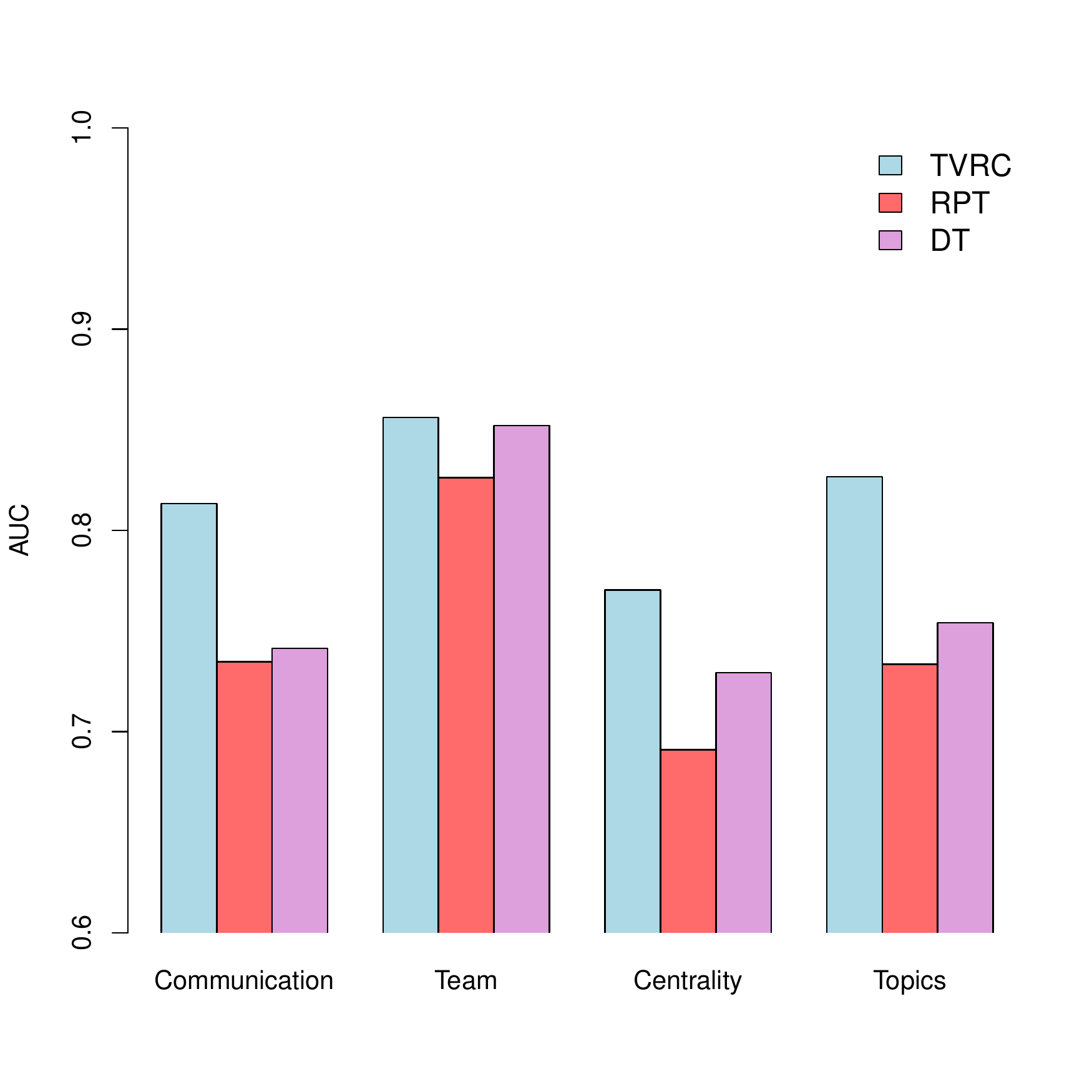}
 \vspace{-4.mm}
\caption{Comparing the utility of the discovered attribute classes and the influence of each with respect to the temporal, relational, and traditional ensembles.}
\label{attr-type-sig}
 \vspace{-6.mm}
\end{figure}

Additionally, the average improvement of the temporal-ensembles is significant at $p<0.05$ with a $16\%$ reduction in error, justifying the proposed temporal ensemble methodologies. From the individual trials, it is clear that the RPT has a lot of variance---despite the use of ensembles, which is aimed at reducing variance, the RPT performs significantly better in one trial ($t=3$) and worse in another ($t=1$). This provides further evidence that relational information and the utility of such information increases significantly when moderated by the temporal-information.

\begin{figure*}[t!]
\centering
 \includegraphics[width=6.5in]{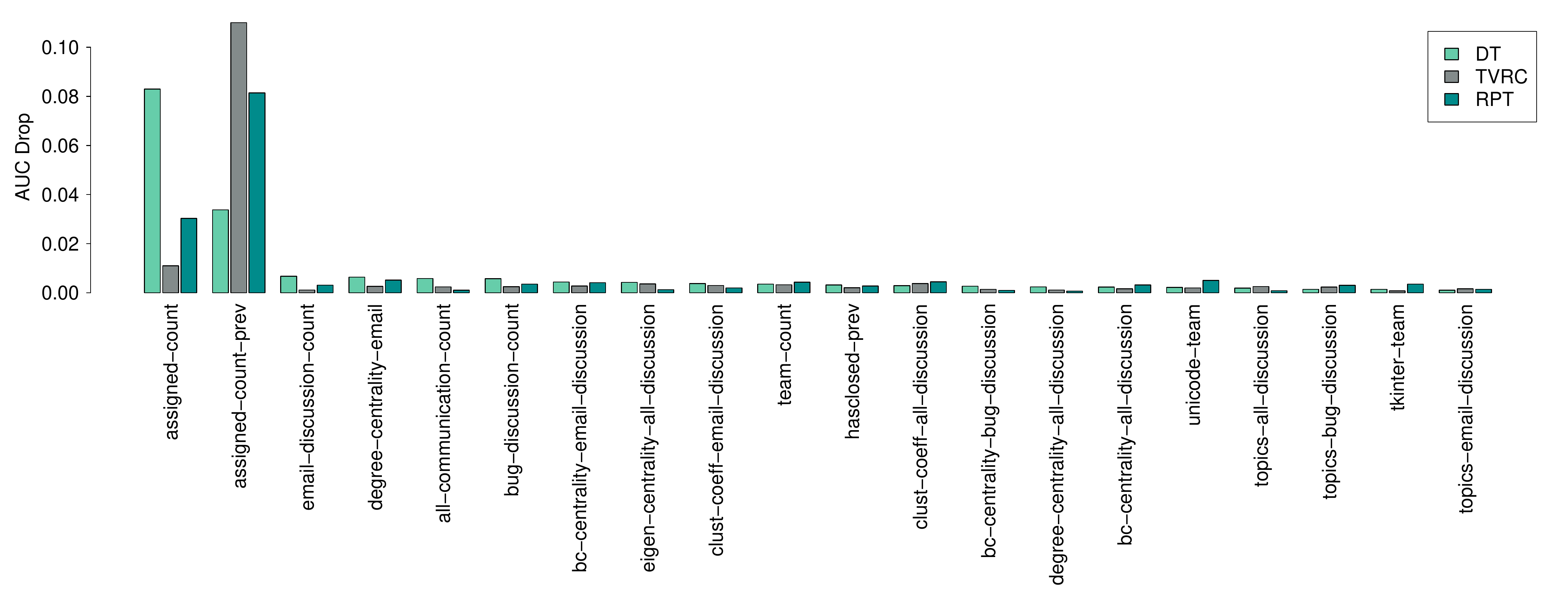}
 \vspace{-5.mm}
\caption{Identifying and ranking of the most significant features in the ensemble models. The significant features used in the \textit{temporal ensemble} are compared to the relational and traditional ensembles. We measure the change in AUC due to the randomization of attribute values.}
\label{rand-results}
 \vspace{-6.mm}
\end{figure*}

\noindent \textbf{Attribute Classes: Temporal Patterns and Significance.} 
We again use one of the most primitive classes of temporal-relational representations in order to tease apart (i.e., more accurately) the most significant attribute category (communication, team, centrality, topics). These primitive temporal-representations also help identify the minimum amount of temporal information that we must consider to outperform relational classifiers. This is important as the more temporal-relational information we exploit, the more complex and expensive it is to learn and search from this space.

In Figure~\ref{attr-type-sig}, we find several striking temporal patterns. First, the team attributes are localized in time and are not changing frequently. For instance, it is unlikely that a developer changes their assigned teams and therefore modeling the temporal dynamics only increases the accuracy by a relatively small percent. However, the temporal-ensemble still increases the accuracy over the other ensemble methods that ignore the temporal patterns. This indicates the robustness of the temporal-relational representations. Moreover, we also notice that a few developers change projects frequently, which could be responsible for the increase in accuracy when the temporal information is leveraged. More importantly, the other classes of attributes are evolving considerably and this fact is captured in the significant improvement of the temporal ensemble models. 
Similar performance is also obtained by varying the temporal granularity. We provide a few examples in the next section.

\noindent \textbf{Randomization.}
We use randomization to identify the significant attributes in the \textit{temporal-ensemble models}. Additionally, randomization provides a means to rank the features and identify redundandt features (i.e., two features may share the same significant temporal pattern).  Randomization is performed on an attribute by randomly reordering the values, thereby preserving the distribution of values but destroying any association of the attribute with the class label. For every attribute, in every time step, we randomize the given attribute, apply the ensemble method, and measure the drop in AUC due to that attribute.
The resulting changes in AUC are used to assess and rank the attributes in terms of their impact on the temporal ensemble (and how it compares to more standard relational or traditional ensembles). Figure \ref{rand-results}. The results are shown in Figure \ref{rand-results}.

We find that the basic traditional ensemble relies more heavily on assignedto (in the current time step) while the temporal ensemble (and even less for the relational ensemble) relies on the previous assignedto attributes. This indicates that the relational information in the past is more useful than the intrinsic information in the present---which points to an interesting hypothesis that a colleagues behavior (and iteractions) precedes their own behavior. Organizations might use this to predict future behavior with less information and proactively respond more quickly.

Additionally, we investigated the attribute classes of each type of ensemble and found that topics are most useful for the temporal ensemble. This indicates that topics are useful as a way to understand the context and strength of interaction among the developers, but only when the temporal dynamics are modeled.

\subsection{Discovering Temporal Patterns}\label{sec:temporal-patterns}
We define three temporal mining techniques based on the temporal framework to construct models with varying temporal granularities. These techniques are combined with relational classifiers or used separately to discover the temporal nature and patterns of relational datasets.

\noindent\textbf{Models of Temporal Granularity.}
If we do not consider temporally weighting the links, nodes, and attributes then we restrict our focus to models based strictly on varying the temporal granularity. In this space, there are a range of interesting models that provide insights into the temporal patterns, structure, and nature of the dataset. We first define three classes of models based on varying the temporal granularity and then evaluate the utility of these models. In addition to discovering temporal patterns, these models are applied to measure the temporal stability and variance of the classifiers over time.

\begin{itemize}
\item \textbf{\textsc{Past}-to-\textsc{Present}.} These  models consider the linked nodes from the distant past and successively increases the size of the window to consider more recent links, attributes, and nodes.
\item \textbf{\textsc{Present}-to-\textsc{Past}.} These models initially consider only the most recent links, nodes, and attributes and successively increase the size of the window to considering more of the past.
\item \textbf{\textsc{Temporal Point}.} These models only consider the links, nodes, and attributes at timestep $k$.
\end{itemize}

\begin{figure}[b!]
 \vspace{-6.mm}
    \subfigure[AI (RBC)]
    	{\label{fig:avg-rbc-topicL1}\includegraphics[width=1.6in]{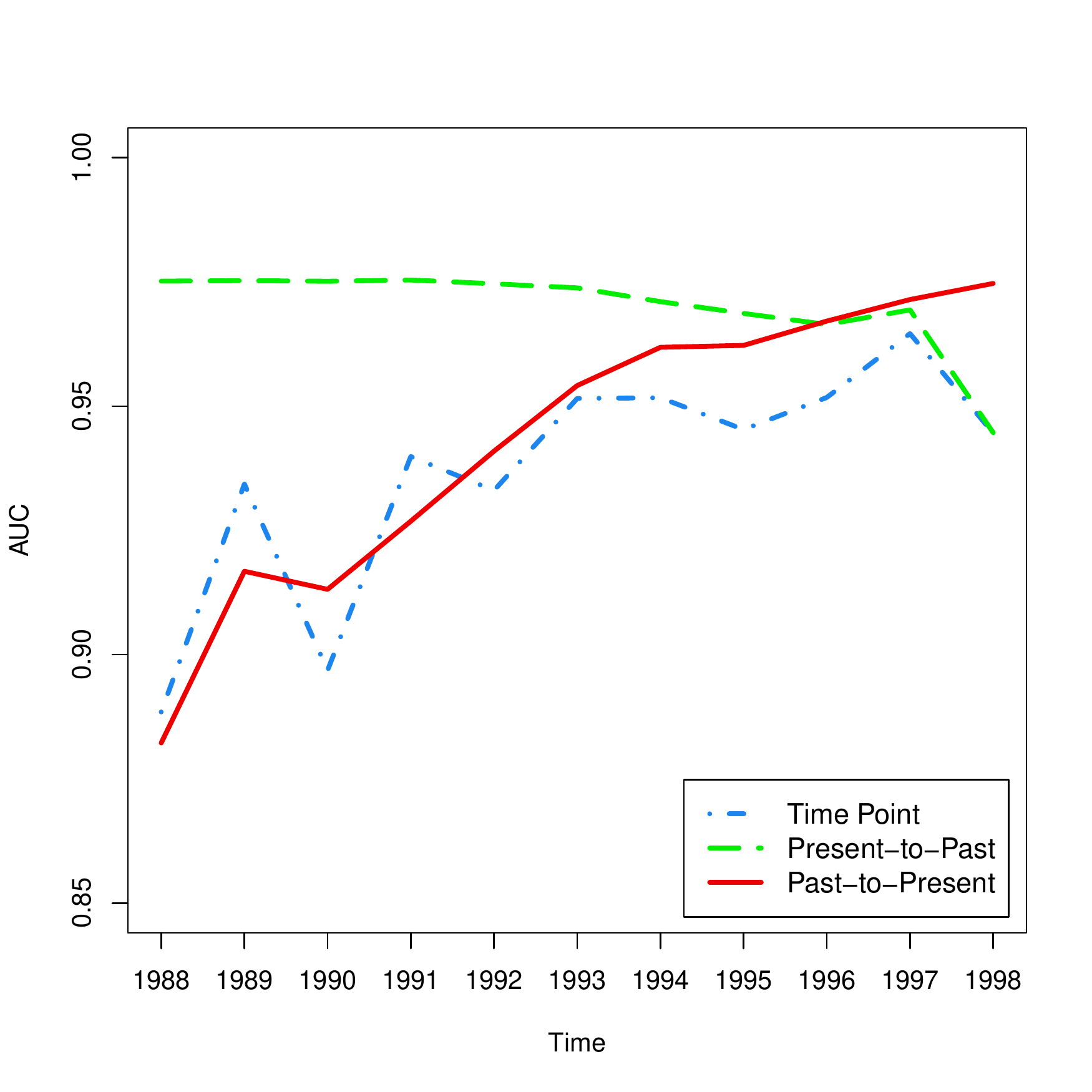}}
    \subfigure[ML (RBC)]
    	{\label{fig:linkprob-topicL1}\includegraphics[width=1.6in]{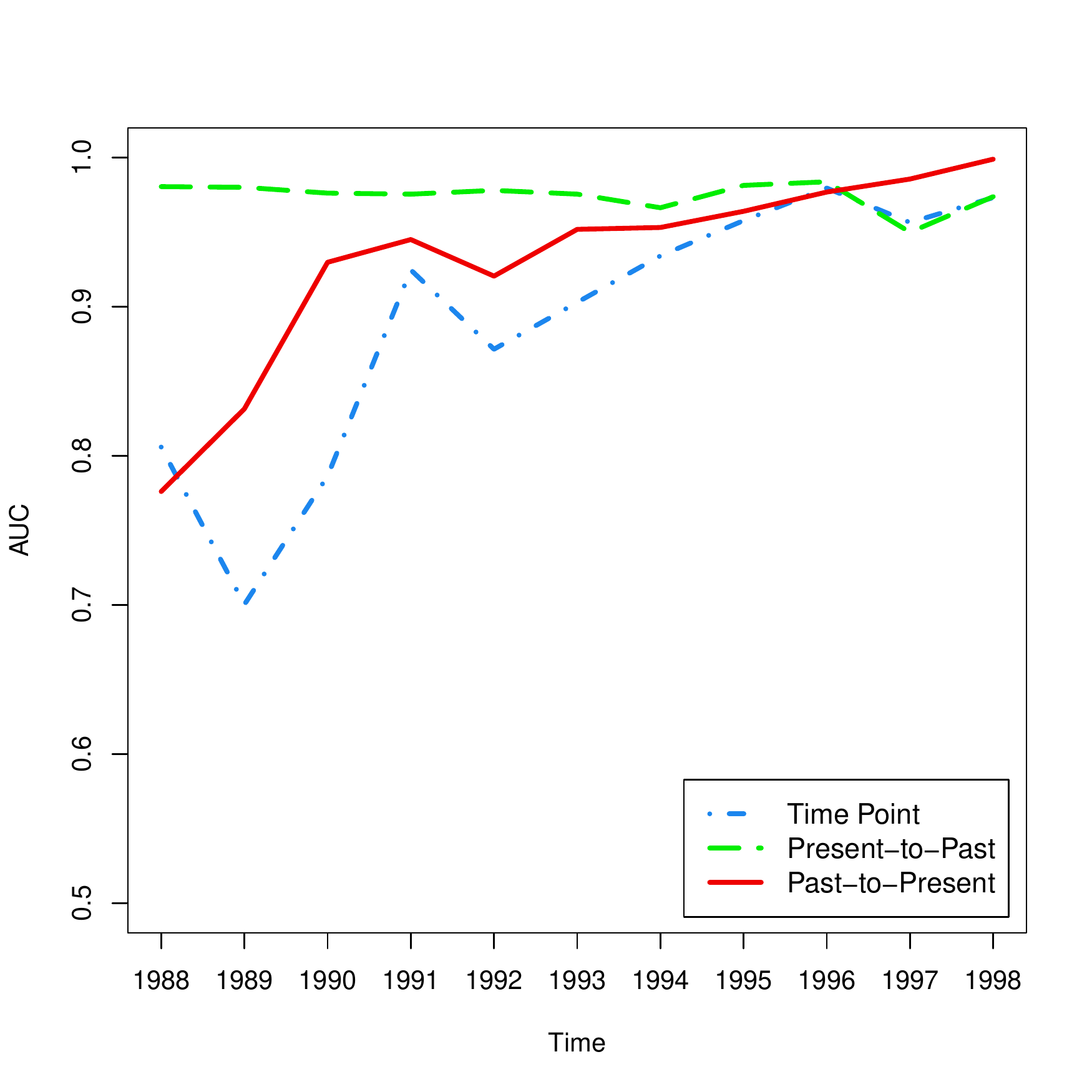}}
  \caption{A variety of temporal granularity models (uniform weighting). Average accuracy using RPT and RBC classifiers for ML and AI prediction tasks.}
  \label{fig:avg-ml}
\end{figure}

\noindent \textbf{Mining Temporal-Relational Patterns}
Intuitively, Figure~\ref{fig:avg-ml} shows that if we consider only the past and successively include more recent information, then the AUC increases as a function of the more recent attributes and links (i.e., \textsc{Past}-To-\textsc{Present} model). 
Conversely, if we consider only the most recent temporal information and successively include more of the past then the AUC initially increases to a local maximum and then dips before increasing as additional past information is modeled. This drop in accuracy indicates a type of temporal-transition in the link structure and attributes. However, we might also expect the values to decay more quickly since papers published in the distant past are generally less \textit{similar} to recent papers as shown previously. Overfitting may justify the slight improvement in AUC as noisey past information is added. The noise reduces bias in training and consequently increases the models ability to generalize for predicting instances in the future. However, this might not always be the case. We expect the noise to be minor in this domain as papers are unlikely to cite other papers in the past that are not related.

More interestingly, the class of \textsc{Temporal-Point} models allow us to more accurately determine if past actions at some previous timestep are predictive of the future and how these behaviors transition over time. These patterns are shown in Figure~\ref{fig:avg-ml}.

\textbf{Temporal Anomalies.} 
The temporal granularity models capture many temporal anomalies. One striking anomaly is seen in Figure~\ref{fig:avg-rbc-topicL1} where the accuracy of the \textsc{Temporal-Point} model decreases significantly in 1990, but then by 1991 the accuracy has increased back to the previous level.

\textbf{Temporal Stability of Relational Classifiers.}
We use the temporal granularity models to compare more accurately the stability of the modified temporal RBC and RPT classifiers which leads us to identify a few striking differences between the two classifiers when modelling temporal networks.

\begin{figure}[!t]
    \subfigure[Temporal Stability (AI)]
    	{\label{fig:avg-classifier-topicL1}\includegraphics[width=1.6in]{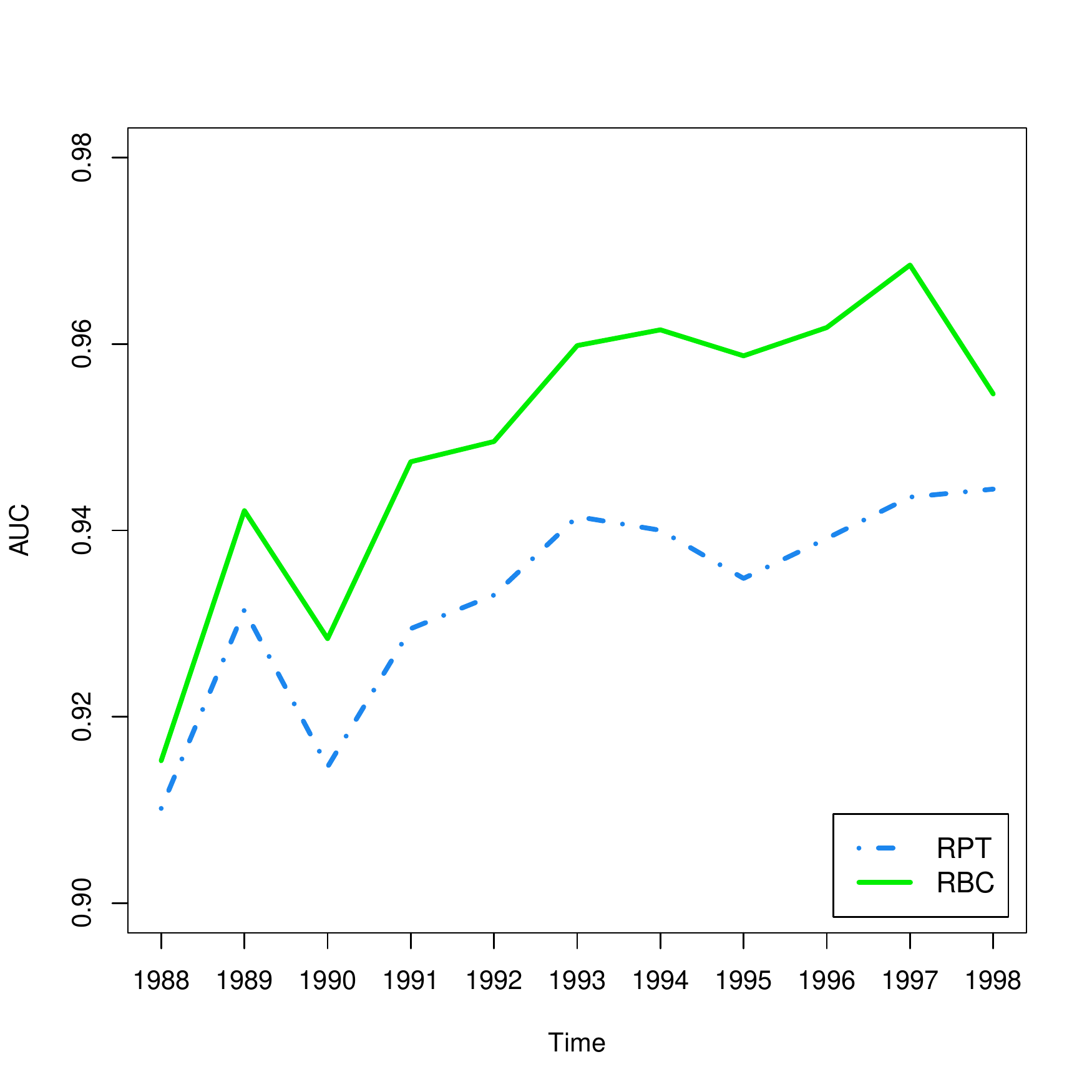}}
    \subfigure[Temporal Stability (ML)]
    	{\label{fig:avg-classifier-topic-ml}\includegraphics[width=1.6in]{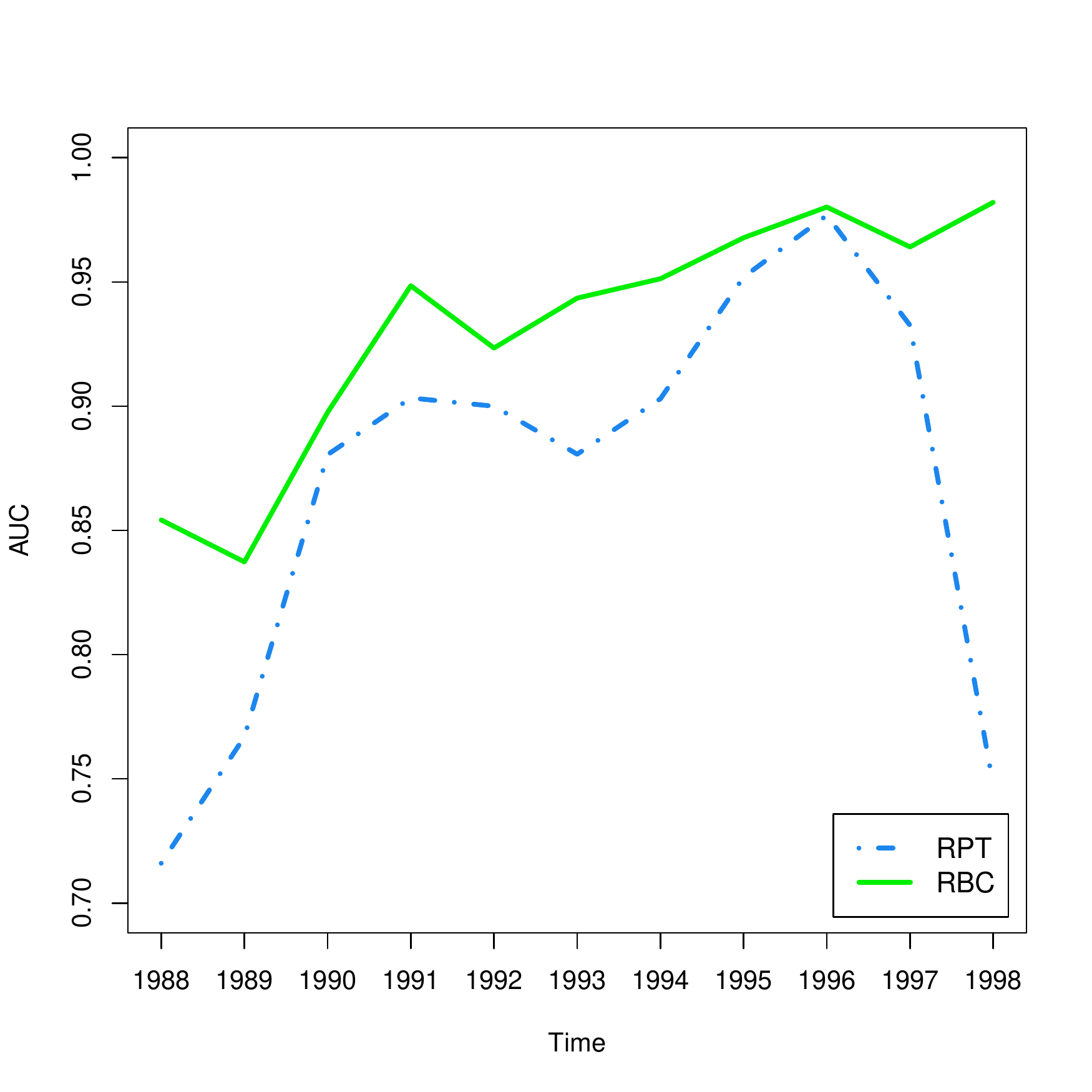}}
  \caption{Average \textit{Temporal Stability} of RPT and RBC for AI and ML prediction tasks.}
  \label{fig:avg-classifier}
 \vspace{-6.mm}
\end{figure}

In Figure~\ref{fig:avg-classifier-topic-ml}, the RBC is shown to be stable over time whereas the variance and stability of the RPT is significantly worse.
This lead us to analyze the internals of the modified RPT and found that for the ML prediction task, the structure of the trees at each timestep are significantly different from one another and consequently unstable. However, we found the structure of the trees to more gradually evolve in the AI prediction task, making the RPT relatively more stable over time.

In addition, we also found the RBC to perform extremely well even with small amounts of temporal information (low support for any hypothesis). 
The RPT and RBC are shown to have complementary advantages and disadvantages, especially for predicting temporal attributes. This provides further justification for the proposed temporal ensemble method that uses both RPT and RBC with each selected temporal-relational representation.

\begin{figure}[b!]
 \vspace{-6.mm}
\subfigure[\textsc{Global Link Recency}]
    	{\label{fig:glr}\includegraphics[width=1.67in]{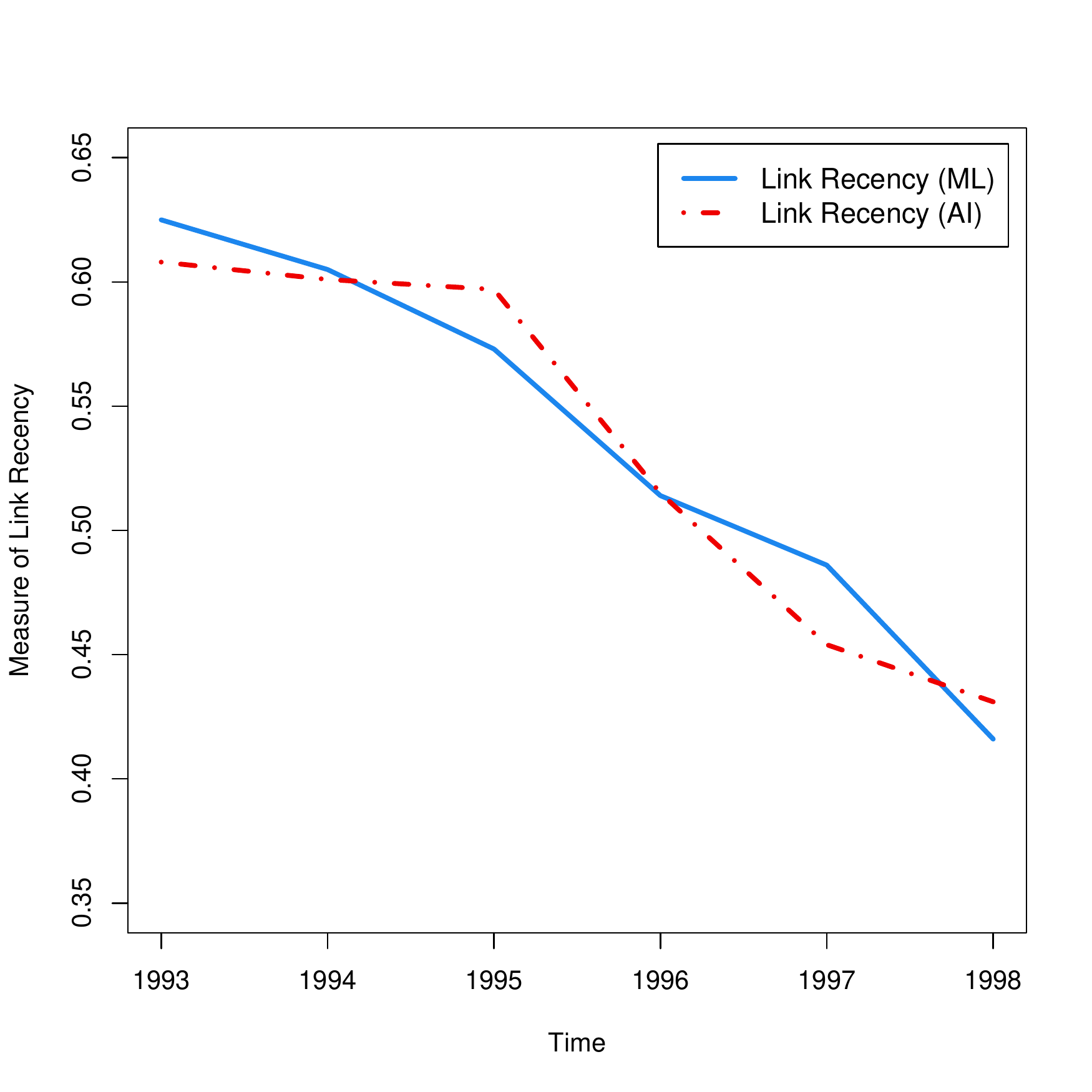}}
\subfigure[\textsc{Temporal Link Probability} (ML)]
    	{\label{fig:linkprob}\includegraphics[width=1.67in]{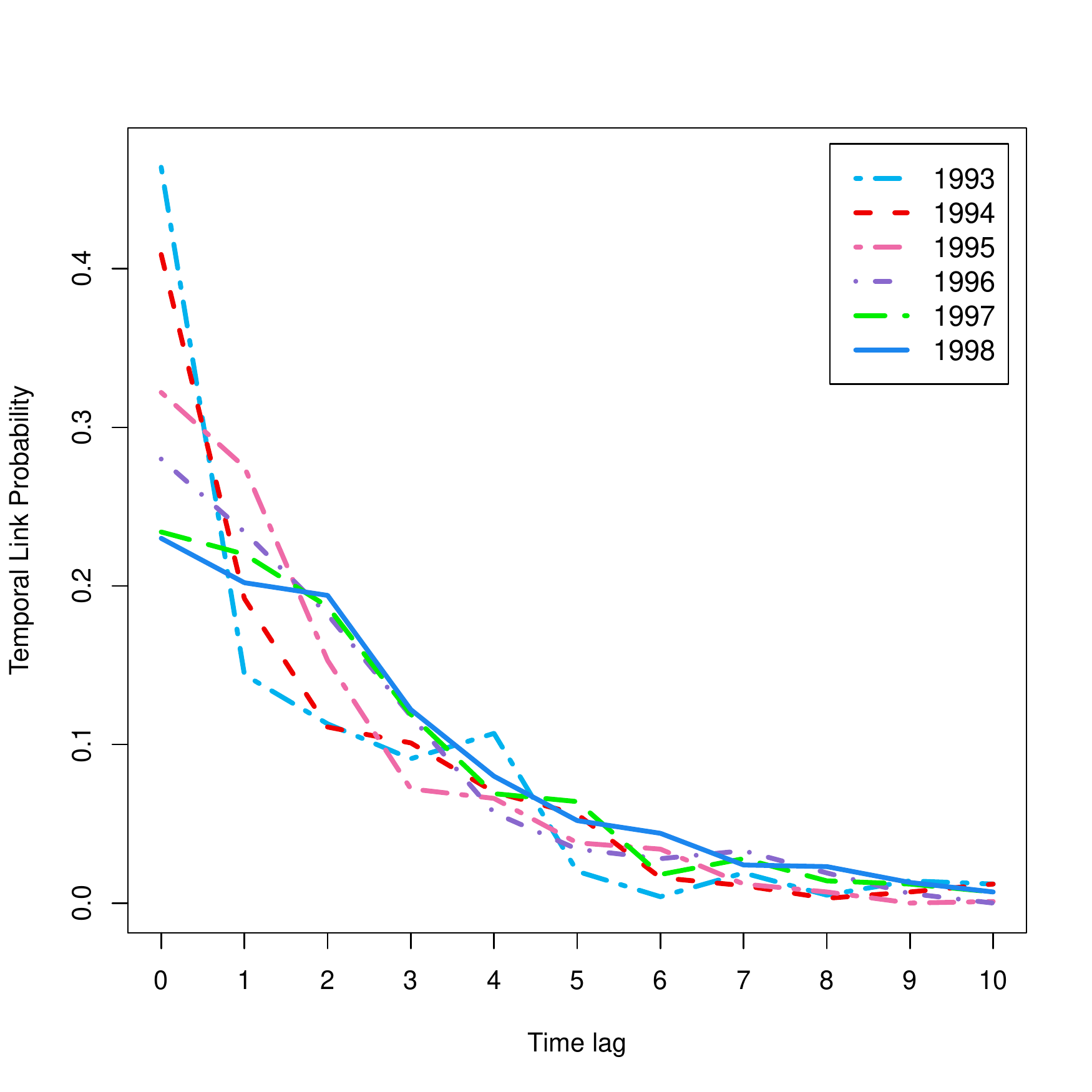}}
  \label{fig:temporal-relational-stats}
\end{figure}

\noindent\textbf{Temporal Relational Statistics.}
The temporal granularity models can be used to compute intuitive yet informative simple measures to gain insights into the temporal nature of a network.
The \textsc{Global Link Recency} measures the probability of citing a paper at time $t$ and $t-1$ for the years 1993-1998 in both AI and ML prediction tasks as shown in Figure~\ref{fig:glr}. For instance, the link recency measure (AI) between 1993 and 1995 is approximately 60\% indicating that out of all the cited papers the majority of them are published in the same year $t$ or the previous year $t-1$. Interestingly, the papers published in the most recent years (e.g., 1998) cite fewer papers from the same year or previous year and more papers in the past (i.e., indicating a temporal-transition that could be due to papers becoming more available to researchers, perhaps with digital archives or other factors). Furthermore, the \textit{temporal relational autocorrelation} measure shows that in general the recent papers are more influential compared to the papers in the past (correlation plots omitted for brevity).

The temporal link probabilities for the AI and ML prediction tasks are shown in Figure~\ref{fig:linkprob}. For the papers in each time period, the probability of citing a paper given the time-lag $\ell$ is computed. Interestingly, the link probabilities at $\ell=3$ for each prediction-time approximately begin to converge. Indicating a global pattern with respect to past links that is independent of the core-nodes initial time period. However, the time-lag between $0 \le \ell \le 3$ captures local patterns with respect to the core-nodes prediction time. Hence, the more recent behavior of the core-nodes is significantly different than their past behavior.

\subsection{Dynamic Textual Analysis: Interpreting Links and Nodes}\label{sec:dynamic-topics}
In this task, we use only the communications to generate a network and then \textit{automatically annotate} the links and nodes by discovering the latent topics of these communications. There are many motivations for such an approach, however, we are most interested in automatically learning evolutionary patterns between the topics and the corresponding developers to increase the accuracy of temporal-relational representations and classifiers.

We first removed a standard list of stopwords and then use a version of Latent Dirichlet Allocation (LDA~\cite{blei:03}) to model the topics over time. We use EM to estimate the parameters and Gibbs sampling for inference. After extracting the latent topics, inference is used to label each link with it's most likely latent topic and each node with their most frequent topic. Instead of this simple representation, we could have used the link probability distributions over time, but found that the potential performance gain did not justify the significant increase in complexity.

The latent topics are modeled in three communication networks (email, bug, and both). From these annotated temporal networks, we investigate the effects of modeling the latent topics of the communications and their evolution over time. We use the discovered evolutionary patterns as features to explore the temporal-relational representations, classifiers, and ensembles and evaluate and compare each of the models.

\begin{table}[!t]
\caption{A set of Discovered Topics and the most significant words}
\label{topics}
\begin{center}
\begin{tabular}{c|c|c|c|c}\multicolumn{1}{c|}{\sc Topic 1} &
\multicolumn{1}{c|}{\sc Topic 2} &
\multicolumn{1}{c|}{\sc Topic 3} &
\multicolumn{1}{c|}{\sc Topic 4} &
\multicolumn{1}{c}{\sc Topic 5}
\\
\hline dev & logged & gt & code & test\\
\hline wrote & patch & file & object & lib\\
\hline guido & issue & lt & class & view\\
\hline import & bugs & line & case & svn\\
\hline code & bug & os & method & trunk\\
\hline pep & problem & import & type & rev\\
\hline mail & fix & print & list & modules\\
\hline release & fixed & call & set & build\\
\hline tests & days & read & objects & amp\\
\hline work & created & socket & change & error\\
\hline people & time & path & imple & usr\\
\hline make & docu & data & functions & include\\
\hline pm & module & error & argument & home\\
\hline ve & docs & open & dict & file\\
\hline support & added & windows & add & run\\
\hline module & check & problem & def & main\\
\hline things & doc & traceback & methods & local\\
\hline good & doesnt & mailto & exception & src\\
\hline van & report & recent & ms & directory\\
\end{tabular}
\end{center}
\vspace{-5.mm}
\end{table}

Table~\ref{topics} lists a few topics and the most significant words for each. We find words with both positive and negative connotation such as `good' or `doesnt'  (i.e., related to sentiment analysis) and also words referring to the domain such as `bugs' or `exception'. Additionally, we find the topics correspond to different development and social aspects. Interestingly, the word `guido' appears significant, since Guido van Rossum is the author of the Python programming language.

\subsection{Modeling the Evolutionary Patterns of Topics}
\begin{figure}[t!]
 \vspace{-8.mm}
\centering
\includegraphics[width=3.4in]{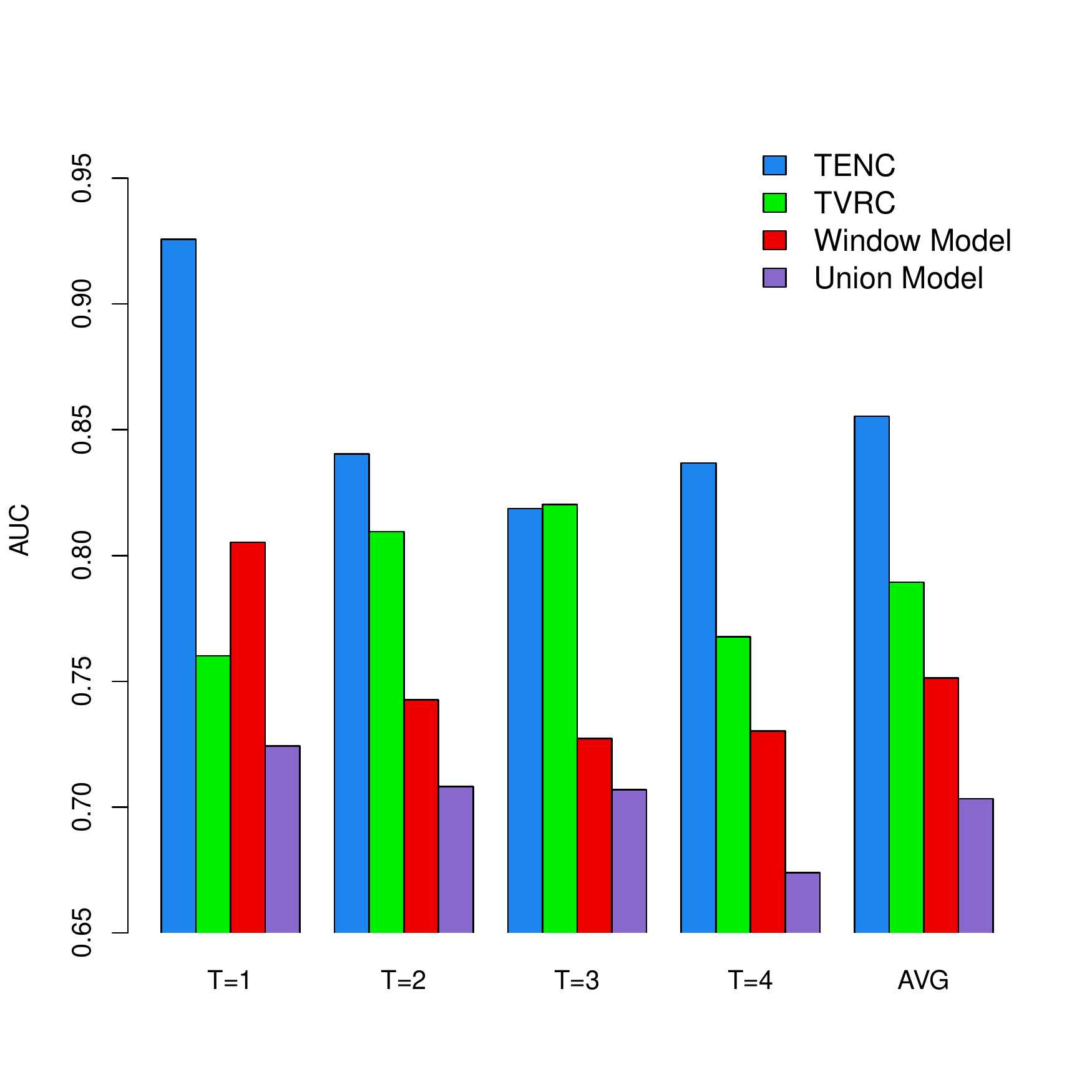}
 \vspace{-12.mm}
\caption{Evaluation of temporal-relational classifiers using only the latent topics of the communications to predict effectiveness. LDA is used to automatically discover the latent topics as well as annotating the communication links and individuals with their appropriate topic in the temporal networks.}
\label{fig:latent-topic-opt}
 \vspace{-6.mm}
\end{figure}

We evaluate these dynamic topic features using various temporal-relational representations for improving classification models. Figure~\ref{fig:latent-topic-opt} indicates the necessity of using a more optimal temporal-relational representation that models the temporal influence of links and attributes. More interestingly, we see that models that consider only simple temporal-relational representations perform significantly worse, indicating that the dynamic topics are only meaningful if appropriately modeled. Additionally, we also learned more complex models from the class of window models to exploit additional temporal granularities, but removed the plots for brevity. In all the experiments, we find that the temporal-relational representations that leverage more of the temporal information outperform models that use only some of the temporal information.

\textit{Evolutionary patterns} between the topics, developers, and their effectiveness are clearly present in annotated networks. These results indicate that productive developers usually communicate about similar topics or aspects of development. Additionally, we find that effective communications have a specific structure that consequently enables others to become more effective. Moreover, these topics and the corresponding communications over time are temporally correlated with a developers effectiveness.

\section{Conclusion}
We proposed a framework for temporal-relational classifiers, ensembles, and more generally, representations for mining temporal data. We evaluate and provide insights of each using real-world networks with different attributes and informational constraints. The results demonstrated the effectiveness, scalability, and flexibility of the temporal-relational representations for classification, ensembles, and mining temporal networks.

\section*{Acknowledgments}
\noindent This research is supported by DARPA and NSF under contract number(s) NBCH1080005 and SES-0823313. This research was also made with Government support under and awarded by DoD, Air Force Office of Scientific Research, National Defense Science and Engineering Graduate (NDSEG) Fellowship, 32 CFR 168a. The U.S. Government is authorized to reproduce and distribute reprints for governmental purposes notwithstanding any copyright notation hereon. The views and conclusions contained herein are those of the authors and should not be interpreted as necessarily representing the official policies or endorsements either expressed or implied, of DARPA, NSF, or the U.S. Government.

\bibliographystyle{IEEEtran}
\bibliography{rossi}

\end{document}